# UzbekStemmer: Development of a Rule-Based Stemming Algorithm for Uzbek Language

Maksud Sharipov [1], Ollabergan Yuldashov [1]

[1] Urgench State University, 14. Kh.Alimdjan str, Urgench City, 220100, Uzbekistan

### Abstract

In this paper we present a rule-based stemming algorithm for the Uzbek language. Uzbek is an agglutinative language, so many words are formed by adding suffixes, and the number of suffixes is also large. For this reason, it is difficult to find a stem of words. The methodology is proposed for doing the stemming of the Uzbek words with an affix stripping approach whereas not including any database of the normal word forms of the Uzbek language. Word affixes are classified into fifteen classes and designed as finite state machines (FSMs) for each class according to morphological rules. We created fifteen FSMs and linked them together to create the Basic FSM. A lexicon of affixes in XML format was created and a stemming application for Uzbek words has been developed based on the FSMs.

## 1. Introduction

Stemming is defined as the conflation of all variations of specific words to a single form called the root or stem. Stemming algorithms for some languages have been published and applied in the building of information retrieval systems, among which for English is the well-known Porter's algorithm [1,2]. According to the stemming principle, stemming algorithms can be divided into four categories, that are rule-based (truncating, affix removal) method, dictionary look-up (table lookup) method, statistical method, and mixed-method [3].

This article presents [4] the application of the stemming task of verbs in the low-resource highly-agglutinative Uzbek language, one of the most important aspects of Natural Language Processing (NLP). The methodology is proposed for doing the stemming of the Uzbek verb words with an affix stripping approach whereas not including any lexicon. Verb affixes are classified into three classes and designed as finite state machines (FSMs) for each class according to morphological rules.

Previous work on stemming algorithms for the Uzbek language [5] were used for normalizing texts. When developing the algorithm, a hybrid approach was used based on the joint application of an algorithmic method, a lexicon of linguistic rules, and a database of normal forms of words in the Uzbek language.

The article [2] offers an intelligent web application developed for the morphological analysis of words in the Uzbek language. The web application is based on the concept of generation and stem analysis of the Uzbek language word forms. A well-known Porter algorithm was chosen as the basis for our stemming approach.

Stemming is an important stage of natural language processing. The research in this paper is a contribution to the development of the stemming tool for the Uzbek language.

Due to the fact that the Uzbek language belongs to the family of agglutinative languages, words are formed as a result of a series of suffixes added to the stem word, which makes it difficult to find the stem of the word [6]. For instance, a single word "O'qi/ma/gan/lar/dan/mi/siz?" (Are you from those who did not read?), which in fact is a question, can be parsed as following:

| | |
|---|---|
| o'qi | to read |
| o'qi/ma | don't read |
| o'qi/ma/gan | he/she di not read |
| o'qi/ma/gan/lar | they did not read |
| o'qi/ma/gan/lar/dan | from those who did not read |
| o'qi/ma/gan/lar/dan/mi/siz | are you from those who did not read |

## 2. Finite State Machines(FSM) Generation

Finite-state technology is very simple and is used a lot for many kinds of tasks in computational linguistics. This technology may be used in phonology, morphology, syntax, spell-checking, tokenization, text-to-speech, data mining, and other fields. The two terms finite-state machines and finite-state automata are synonymous. A finite-state machine is a machine consisting of a set of states with a set of allowable inputs which change the state as well as a set of outputs [7,8].

We use FSMs to create the stemming algorithm, which allows us to analyze words from right to left. There are four stages to create these FSMs [9]:

- Creating a left to right FSM;
- Labeling the suffix   es
- Inverting the left to right FSM and obtaining a non deterministic finite state automaton (NFA)
- Converting NFA to a deterministic finite automaton (DFA) and constructing the right to left FSM

Affixes were divided into 15 classes according to FSM and were stored in XML format.

**Table 1**
Affix classes

| Class | Class name | Type | Affixes | Allomorphs |
|---|---|---|---|---|
| 1 | Particle suffixes | Inflectional | 10 | 12 |
| 2 | Declension suffixes | Inflectional | 21 | 29 |
| 3 | Conjugation suffixes | Inflectional | 34 | 15 |
| 4 | Participle & Gerund suffixes | Inflectional | 12 | 19 |
| 5 | Verbal Adverb suffixes | Inflectional | 10 | 20 |
| 6 | Relative verb suffixes | Inflectional | 15 | 25 |
| 7 | Derivational [from verb to noun] | Derivational | 8 | 13 |
| 8 | Noun & Adjective suffixes | Inflectional | 11 | 12 |
| 9 | Number suffixes | Inflectional | 11 | 12 |
| 10 | Pronouns suffixes | Inflectional | 2 | 2 |
| 11 | Prefixes | Derivational | 7 | 7 |
| 12 | Derivational [Verb] suffixes | Derivational | 67 | 73 |
| 13 | Derivational [Adjective] suffixes | Derivational | 113 | 137 |
| 14 | Derivational [Noun] suffixes | Derivational | 106 | 124 |
| 15 | Derivational [Adverb] suffixes | Derivational | 27 | 33 |
| | | **Total** | 454 | 533 |

An affix in Uzbek can have multiple allomorphs in order to provide sound harmony (as the phonological rules) in the word to which it is affixed. For example, the Declension suffix with generic representation –*Ga* has three allomorphs: –*ga*, –*ka*, –*qa*. The abbreviations used to show suffixes in a generic way are shown below:

G: g, k, q        Y: a, y        K: k, g        Q: k, g, g', q        T: t, d        A: a,o
(): the letter between parentheses can be omitted

All affixes are stored in the Suffixes.xml file. The structure of the XML file is as follows:

```xml
<?xml version='1.0' encoding='UTF-8'?>
 <suffixes>
      <item fsm_id="" suff_id=""  group="" pos="" class="">
              <suffix allomorph="false">
                      <name>...</name>
                      <exception_cut>...</exception_cut>
                      <exception_pass>...</exception_pass>
              </suffix>
              <definition>...</definition>
              <old_pos>...</old_pos>
              <new_pos>...</new_pos>
      </item>
 </suffixes>
```

Here, the <suffixes> ... </suffixes> tag is the root tag, <item>...</item> teg for affixes, there are fsm_id(for  FSM  id in Table 1), suff_id(for affixes id in each FSM Suffixes table), group="" category="" class="" type="" ( for to express the internal division of affixes in Table 2 within word classes) attributes in <item>...</item>. The <exception_cut> </exception_cut> and <exception_pass> </exception_pass> tags are used for exceptions. The <old_pos> </old_pos> and <new_pos> </new_pos> tags indicate the state before and after the affixes under consideration in the context of open word classes. If the allomorph attribute of the <suffixes> ... </suffixes> tag is true, all allomorphs of the affixes are entered as:

```xml
<suffix allomorph="true">
        <allomorph>
                <name>...</name>
                <exception_cut>...</exception_cut>
                <exception_pass>...</exception_pass>
        </allomorph>
        <allomorph>
                <name>...</name>
                <exception_cut>...</exception_cut>
                <exception_pass>...</exception_pass>
        </allomorph>
</suffix>
```

## 2.1.  Declension Suffixes

Declension suffixes are added in Uzbek after open word classes such as noun, adjective, pronoun, numeral and past participle. First, we learn the sequence of affixes, then we build the FSM from left to right [6,10,11,12,13].

In Figure 1, a numerical value is indicated a state. In the following stages, the states will be expressed with these numbers: 0 – ending state, 1 – initial state. The character "ε" means the empty transitions between the states. The final state is represented by 2 circles.

In the next step, the affixes in Figure 1 are numbered. In the next steps, the sequence number in Table 2 is used instead of the affixes.

In Figure 2, the right to left NFA is shown, in which the FSM position from left to right in reverse order is constructed from right to left.

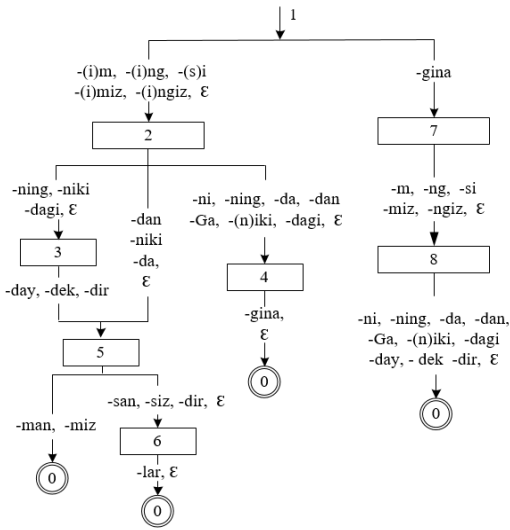

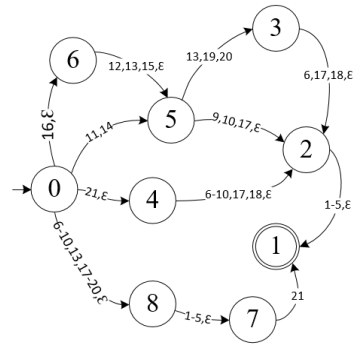

**Figure 1**: Declension Suffixes left to right FSM          **Figure 2**: Declension Suffixes right to left NFA

Let us look at some examples of how the following Declension suffixes are made from left to right finite automata (Figure 1): qishlog' (1) → im (2) → dan (5) → siz (6) → lar (0) (You are from my village), where the numbers in parentheses indicate the transitions in Figure 1.

The large number of gaps in the FSM from left to right and the depth of the FSM complicate the construction of the FSM from right to left.

Table 3 illustrates the transition from NFA to DFA. In DFA, there is only one way out of each entry and exit. There will be no empty transaction at DFA.

**Table 2**

Declension Suffixes

| | | | | | |
|---|---|---|---|---|---|
| 1. | –(i)m | 8. | –Ga | 15. | –siz |
| 2. | –(i)ng | 9. | –da | 16. | –lar |
| 3. | –s(i) | 10. | –dan | 17. | –(n)iki |
| 4. | –(i)miz | 11. | –man | 18. | –dagi |
| 5. | –(i)ngiz | 12. | –san | 19. | –day |
| 6. | –ning | 13. | –dir | 20. | –dek |
| 7. | –ni | 14. | –miz | 21. | –gina |

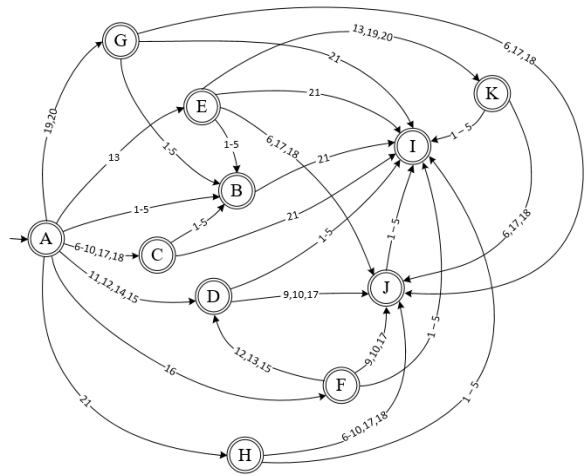

**Figure 3**: Declension Suffixes right to left FSM

If we define the input state of the DFA as A, it includes the input state of the NFA 0 and the states 1,2,4,5,6,7,8 that can pass through the 0-statement empty transition. Entry in DFA A state includes all {0,1,2,4,5,6,7,8} initial cases in NFA. We analyze the new states formed as a result of trimming the suffixes of each state, after the initial case. For example, if the initial state of a word in state A is defined by the suffixes "1-5"(1 - (i)m, 2 - (i)ng, 3 - (s)i, 4 - (i)miz, 5 - (i)ngiz) in Table 2, it will cut the suffix defined on the right side of the word and move the word to the next state B.

When depicting the FSM from right to left, the state contains the final state of the NFA - 1, while from right to left the state of the FSM is the final state and is represented by 2 circular circles. In this case, the FSM ends in the state if no other suffix is found after the word that cut the suffix and the incoming state is the final state.

**Table 3**
Declension Suffixes NFA to DFA Operation

| A={0,1,2,4,5,6,7,8} | | "13,19,20" : | T={3} →{1,2,3} →K |
|---|---|---|---|
| "1-5" : | T={1,7} →B | "6,17,18" : | T={2} →{1,2} →J |
| "6-10,17,18" : | T={2,8} →{1,2,7,8} →C | **F={1,2,5,6}** | |
| "11,12,14,15" : | T={5} →{1,2,5} →D | "1-5" : | T={1} →I |
| "13" : | T={3,5,8} →{1,2,3,5,7,8} →E | "9,10,17" : | T={2} →{1,2} →J |
| "16" : | T={6} →{1,2,5,6} →F | "12,13,15" : | T={5} →{1,2,5} →D |
| "19,20" : | T={3,8} →{1,2,3,7,8} →G | **G={1,2,3,7,8}** | |
| "21" : | T={1,4} →{1,2,4} →H | "1-5" : | T={1,7} →B |
| **B={1,7}** | | "21" : | T={1} →I |
| "21" : | T={1} →I | "6,17,18" : | T={2} →{1,2} →J |
| **C={1,2,7,8}** | | **H={1,2,4}** | |
| "1-5" : | T={1,7} →B | "1-5" : | T={1} →I |
| "21" : | T={1} →I | "6-10,17,18" : | T={2} →{1,2} →J |
| **D={1,2,5}** | | **J={1,2}** | |
| "1-5" : | T={1} →I | "1-5" : | T={1} →I |
| "9,10,17" : | T={1,2} →J | **K={1,2,3}** | |
| **E={1,2,3,5,7,8}** | | "1-5" : | T={1} →I |
| "1-5" : | T={1,7} →B | "6,17,18" : | T={2} →{1,2} →J |
| "21" : | T={1} →I | | |

## 2.2. Noun & Adjective Suffixes

Figure 4 shows the FSM of the Noun & Adjective suffixes from left to right. This is an integral continuation of the Declension additions. This FSM has a relatively simple structure.

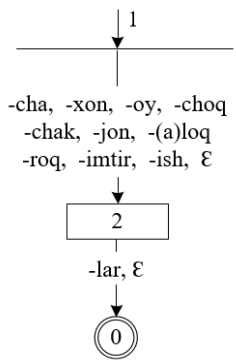

**Figure 4**: Noun & Adjective Suffixes left to right FSM

**Table 4**
Noun & Adjective Suffixes

| 1. | –cha | 7. | –(a)loq |
|---|---|---|---|
| 2. | –xon | 8. | –roq |
| 3. | –oy | 9. | –imtir |
| 4. | –choq | 10. | –ish |
| 5. | –chak | 11. | –lar |
| 6. | –jon | | |

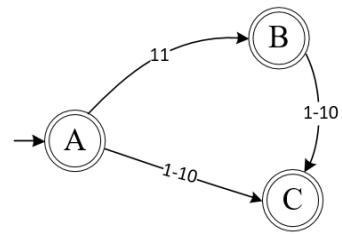

**Figure 5**: Noun & Adjective Suffixes right to left FSM

If we analyze the word "*yaxshiroqlaridan*" *(from things/people which are better)*. The word initially passes in Figure 3, through Declension Suffixes FSM from right to left, and in this FSM, 10(dan) suffix of the word are cut and go from A to C state, then FSM cut the 3((s)i) suffix and it move from C to B state: "*yaxshiroqlaridan*"*( A− 10 → C ) → "yaxshiroqlari"( C − 3 →B)*. It stops here because no suffixes after B state are found and B state is the final state. The resulting "*yaxshiroqlar*"*(things/people which are better)* word is passed to the next Noun&Adjective Suffixes FSM in Figure 5. In this FSM, the suffix 11(lar) is cut and moves from state A to state B, then cut the suffix 8 (roq) and move from B to C state: "*yaxshiroqlar*"*( A − 11 → B) → "yaxshiroq" ( B − 8 → C) → "yaxshi" (good)*. The result is the stem of the word.

## 2.3. Derivational [Noun] Suffixes

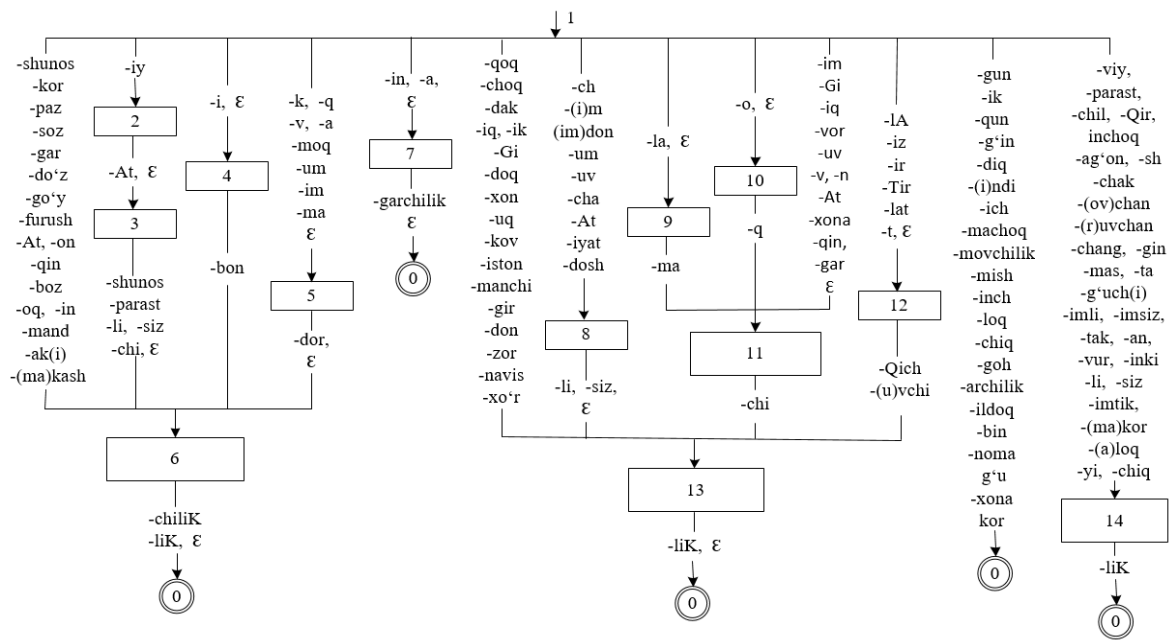

**Figure 6**: Derivational [Noun] Suffixes left to right FSM

Most suffixes in Uzbek belong to the category of nouns[14]. Figure 6 depicts the left to right Noun Derivational Suffixes. In Uzbek, Derivational suffixes belonging to several different word groups can be added to the same base. In order to reduce errors in Figure 6, other word group suffixes that are added to the base before Noun Suffixes are also given without spaces.

**Table 5**
Derivational [Noun] Suffixes

| | | | | | | | | | |
|---|---|---|---|---|---|---|---|---|---|
| 1. | −shunos | 23. | −i | 45. | −manchi | 67. | −(i)ndi | 89. | −(r)uvchan |
| 2. | −(ma)kor | 24. | −bon | 46. | −gir | 68. | −ich | 90. | −chang |
| 3. | −paz | 25. | −k | 47. | −don | 69. | −machoq | 91. | −gin |
| 4. | −soz | 26. | −q | 48. | −zor | 70. | −movchilik | 92. | −mas |
| 5. | −gar | 27. | −v | 49. | −navis | 71. | −mish | 93. | −ta |
| 6. | −do'z | 28. | −moq | 50. | −xo'r | 72. | −inch | 94. | −g'uch(i) |
| 7. | −go'y | 29. | −a | 51. | −ch | 73. | −loq | 95. | −imli |
| 8. | −furush | 30. | −um | 52. | −(i)m | 74. | −chiq | 96. | −imsiz |
| 9. | −At | 31. | −im | 53. | −imdon | 75. | −goh | 97. | −tak |
| 10. | −on | 32. | −ma | 54. | −uv | 76. | −archilik | 98. | −an |
| 11. | −qin | 33. | −dor | 55. | −cha | 77. | −ildoq | 99. | −vur |
| 12. | −boz | 34. | −qoq | 56. | −iyat | 78. | −bin | 100. | −inki |
| 13. | −oq | 35. | −choq | 57. | −dosh | 79. | −noma | 101. | −imtik |
| 14. | −in | 36. | −dak | 58. | −la | 80. | −g'u | 102. | −(a)loq |
| 15. | −mand | 37. | −iq | 59. | −o | 81. | −viy | 103. | −yi |
| 16. | −ak(i) | 38. | −ik | 60. | −vor | 82. | −chil | 104. | −liK |
| 17. | −(ma)kash | 39. | −Gi | 61. | −n | 83. | −Qir | 105. | −chiliK |
| 18. | −iy | 40. | −doq | 62. | −xona | 84. | −inchoq | 106. | −garchiliK |
| 19. | −parast | 41. | −xon | 63. | −gun | 85. | −ag'on | | |
| 20. | −li | 42. | −uq | 64. | −qun | 86. | −sh | | |
| 21. | −siz | 43. | −kov | 65. | −g'in | 87. | −chak | | |
| 22. | −chi | 44. | −iston | 66. | −diq | 88. | −(ov)chan | | |

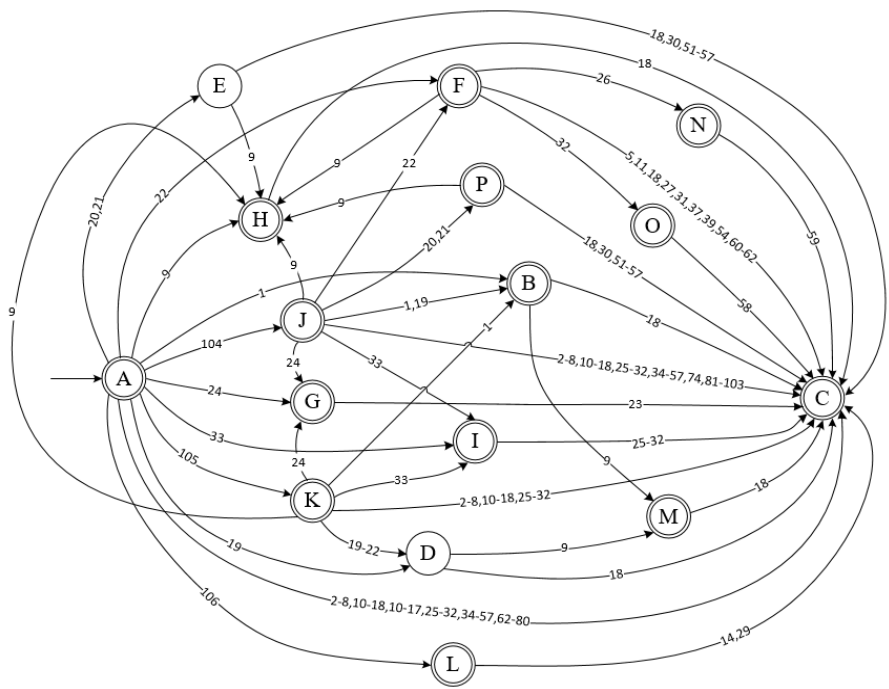

**Figure 7**: Derivational [Noun] Suffixes right to left FSM

## 2.4. Main FSM

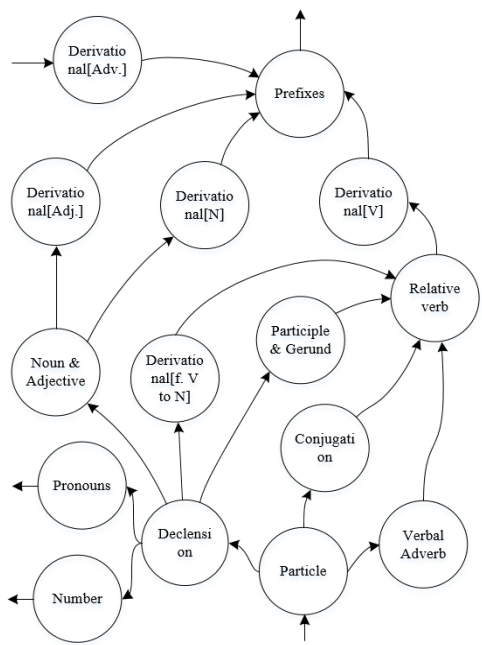

**Figure 8**: Final FSM

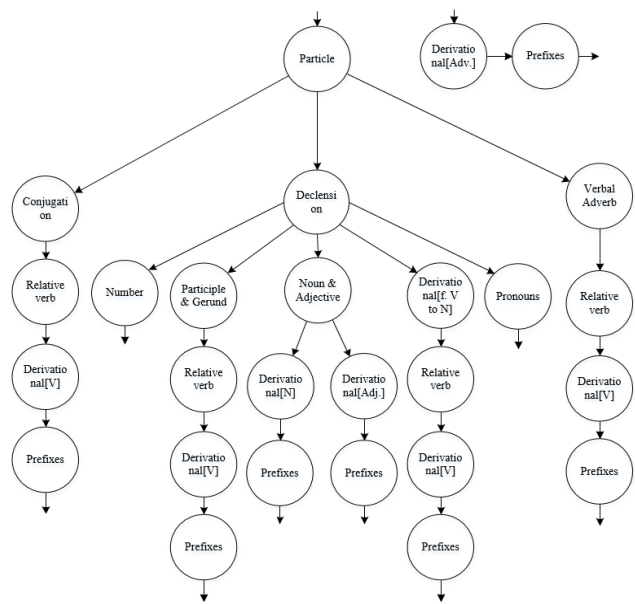

**Figure 9**: Final FSM Tree

The main FSM has entry from 2 FSMs: Particle and Derivational [Adv]. If adverb affixes are added to the stem, inflectional ones are not added. A sequence of connections of FSMs is given in the main FSM. Each word should be in the entry part of the main FSM. Words are passed from one FSM to another. There are 3 outputs in the main FSM: Prefixes, Pronouns, and Numbers. There are 9 paths in the main FSM. The word is passed through all the ways whichever way many affixes are stripped, we take the word in this way as a stem. Depending on the way in which the stem is found, it is possible to tell which word classes it belongs to.

## 3. Conclusion & Future Work

Currently, the issue of creating an Uzbek stemmer has not been resolved due to insufficient NLP resources for the Uzbek language. Therefore, it is developed for Uzbek by using the rule-based structure of the language. In Uzbek sentences, words are usually concatenated with lots of suffixes, that is the reason an information retrieval system needs a quick performing stemming algorithm. In systems that work with root detection, it is recommended not to use a dictionary to increase application speed. Therefore, this study does not use any word databases. The considered method in this study can be the basis for the development of an Uzbek stemming algorithm and according to the approaches (algorithm) the python library developed that is easy to install and use in NLP applications. The software and instructions can be reached from the address [15]. This will be applied to develop a software in information retrieval system for Uzbek documents.

## 4. References

[1]    Porter MF. An algorithm for suffix stripping. 1980; 130–137.
[2]    Mengliev D, Barakhnin V, Abdurakhmonova N. Development of intellectual web system for morph analyzing of uzbek words. Applied Sciences (Switzerland) 2021; 11.
[3]    Tengku Mohd T. Sembok, Belal Mustafa Abu Ata, Zainab Abu Bakar. A Rule and Template Based Stemming Algorithm for Arabic Language. 2011 974–981
[4]    Maksud Sharipov, Ulugbek Salaev, Gayrat Matlatipov. IMPLEMENTED STEMMING ALGORITHMS BASED ON FINITE STATE MACHINE FOR UZBEK VERBS | COMPUTER LINGUISTICS: PROBLEMS, SOLUTIONS, PROSPECTS. 2021
[5]    Bakayev I. View of DEVELOPMENT OF A STEMMING ALGORITHM BASED ON A LINGUISTIC APPROACH FOR WORDS OF THE UZBEK LANGUAGE. 2021 195–202
[6]    Sapayev Qalandar. Hozirgi o'zbek tili(morfemika, so'z yasalishi va morfologiya). 2009.
[7]    Tayebeh Mosavi Miangah. Finite-State Technology in Natural Language Processing. 2014
[8]    Bakaev I. Creation of a morphological analyzer based on finite-state techniques for the Uzbek language. 2020.
[9]    Eryiğit G, Adalı E. AN AFFIX STRIPPING MORPHOLOGICAL ANALYZER FOR TURKISH. 2004; 299–304.
[10]   Hamroyeva Orzigul. Ona tili ma'ruzalar to'plami. 2021;
[11]   Yormat Tojiyev. O'zbek tili morfemikasi. 1992.
[12]   M.Hamroyev, D.Muhamedova, va b. Ona tili. Toshkent: IQTISOD-MOLIYA, 2007.
[13]   Tojiyev Yormat, Nazarova Nodira, Tojiyeva Gulchexra. O'zbek tilidagi ergash morfemalarning semantik-stilistik xususiyatlari. Toshkent, 2012.
[14]   Azim Hojiev. O'zbek tili morfologiyasi, morfemikasi va so'z yasalishining nazariy masalalari. Toshkent, 2010.
[15]   Maksud Sharipov, Salaev Ulugbek, Ollabergan Yuldashov, Sobirov Jasurbek. Stemming Algorithm for Uzbek language: UzbekStemmer. 2021 https://www.higithub.com/MaksudSharipov/repo/UzbekStemmer